\documentclass[sigconf]{acmart}
\AtBeginDocument{%
  }

\setcopyright{acmlicensed}
\copyrightyear{2026}
\acmYear{2026}
\acmDOI{XXXXXXX.XXXXXXX}
\acmConference[MM '26]{the 34rd ACM International Conference on Multimedia}{November 10--14,
  2026}{Rio de Janeiro, Brazil}
\usepackage{cleveref}
\crefname{table}{Table}{Tables}
\crefname{figure}{Figure}{Figures}
\usepackage{colortbl}
\usepackage{array}
\usepackage{amsmath, bm}
\usepackage{multirow}
\usepackage{makecell}
\usepackage{graphicx}
\usepackage{algorithmicx,algorithm}
\usepackage[dvipsnames,svgnames]{xcolor}
\usepackage{xcolor}
\usepackage{svg}
\usepackage{alltt}
\usepackage{caption}
\usepackage{wrapfig}
\usepackage{pifont}       % \ding{xx}
\usepackage{bbding}       % \Checkmark,\XSolid,... (需要和pifont宏包共同使用)
\usepackage{fontawesome}  % \faCheck,\faTimes
\usepackage{tikz}
\usepackage{booktabs}

\definecolor{pink}{HTML}{DAA2B6}
\definecolor{brownnew}{HTML}{AA5757}
\definecolor{greennew}{HTML}{91CD68} 
\definecolor{bluenew}{HTML}{5EA8ED}
\definecolor{orangenew}{HTML}{E6AC83}

\usepackage[table]{xcolor} % 支持表格上色
\definecolor{lightblue}{RGB}{218,227,243} % 浅蓝色 (RGB)
\setcopyright{acmlicensed}
\copyrightyear{2026}
\acmYear{2026}
\acmDOI{XXXXXXX.XXXXXXX}% 录用后再填
\acmISBN{978-1-4503-XXXX-X/2018/06}
\acmSubmissionID{1232}   % 替换成你的 Submission ID

\begin{document}

\title{HiMix: Hierarchical Artifact-aware Mixup for Generalized Synthetic Image Detection}

\author{Shuchang Zhou}
\affiliation{%
  \institution{University of Electronic Science and Technology of China}
  \city{Chengdu, Sichuan}
  \country{China}}
\email{sczhou@std.uestc.edu.cn}

\author{Kaiwen Shen}
\affiliation{%
  \institution{University of Electronic Science and Technology of China}
  \city{Chengdu, Sichuan}
  \country{China}}
\email{ukevin.shen@std.uestc.edu.cn}

\author{Jiwei Wei}
\authornote{Corresponding author.}
\affiliation{%
  \institution{University of Electronic Science and Technology of China}
  \city{Chengdu, Sichuan}
  \country{China}}
\email{mathematic6@gmail.com}

\author{Yuyang Zhou}
\affiliation{%
  \institution{Hainan University}
  \city{Haikou, Hainan}
  \country{China}}
\email{yuyangzhouzhou@hainanu.edu.cn}

\author{Peng Wang}
\affiliation{%
  \institution{University of Electronic Science and Technology of China}
  \city{Chengdu, Sichuan}
  \country{China}}
\email{wangpeng8619@gmail.com}

\author{Yang Yang}
\affiliation{%
  \institution{University of Electronic Science and Technology of China}
  \city{Chengdu, Sichuan}
  \country{China}}
\email{yang.yang@uestc.edu.cn}


\begin{abstract}
  The rapid evolution of generative models has enabled the creation of highly realistic and diverse synthetic images, posing significant challenges to reliable and generalizable Synthetic Image Detection (SID). However, existing detectors are typically trained on limited and biased datasets, resulting in poor generalization to unseen generators. To address this issue, we propose HiMix, a unified framework that enhances generalization by expanding the training distribution and promoting artifact-aware representations. Specifically, the Mixup-driven Distributional Augmentation (MDA) module constructs continuous transitional samples between real and fake images, improving coverage of low-confidence regions and exposing the model to more challenging samples, while the pixel-wise mixup operation smoothly perturbs semantics to enhance sensitivity to low-level artifacts. Moreover, the Hierarchical Artifact-aware Representation (HAR) module aggregates artifact information from both global and local levels through cross-layer integration and coarse-to-fine feature fusion, enabling the extraction of discriminative forgery representations under diverse distributions. Extensive experiments across multiple benchmarks demonstrate that HiMix achieves state-of-the-art performance, establishing well-separated logits for improved generalization to unseen forgeries.
\end{abstract}





\maketitle

\section{Introduction}
Recently, advances in generative models such as Generative Adversarial Networks (GANs)~\cite{goodfellow2014generative,karras2020analyzing} and Diffusion Models (DMs)~\cite{kim2022diffusionclip,ho2020denoising} have enabled the low-cost creation of high-quality images, but also raised concerns about misinformation, identity forgery, privacy leakage, and financial fraud~\cite{bontridder2021role,tolosana2020deepfakes,golda2024privacy}. As synthetic images become increasingly realistic and diverse, Synthetic Image Detection (SID), which aims to distinguish natural from synthetic images, has become an essential task. Developing SID methods that are both reliable and generalizable remains a key challenge for research and real-world deployment.

\begin{figure*}[t]
  \centering
  \includegraphics[width=\textwidth]{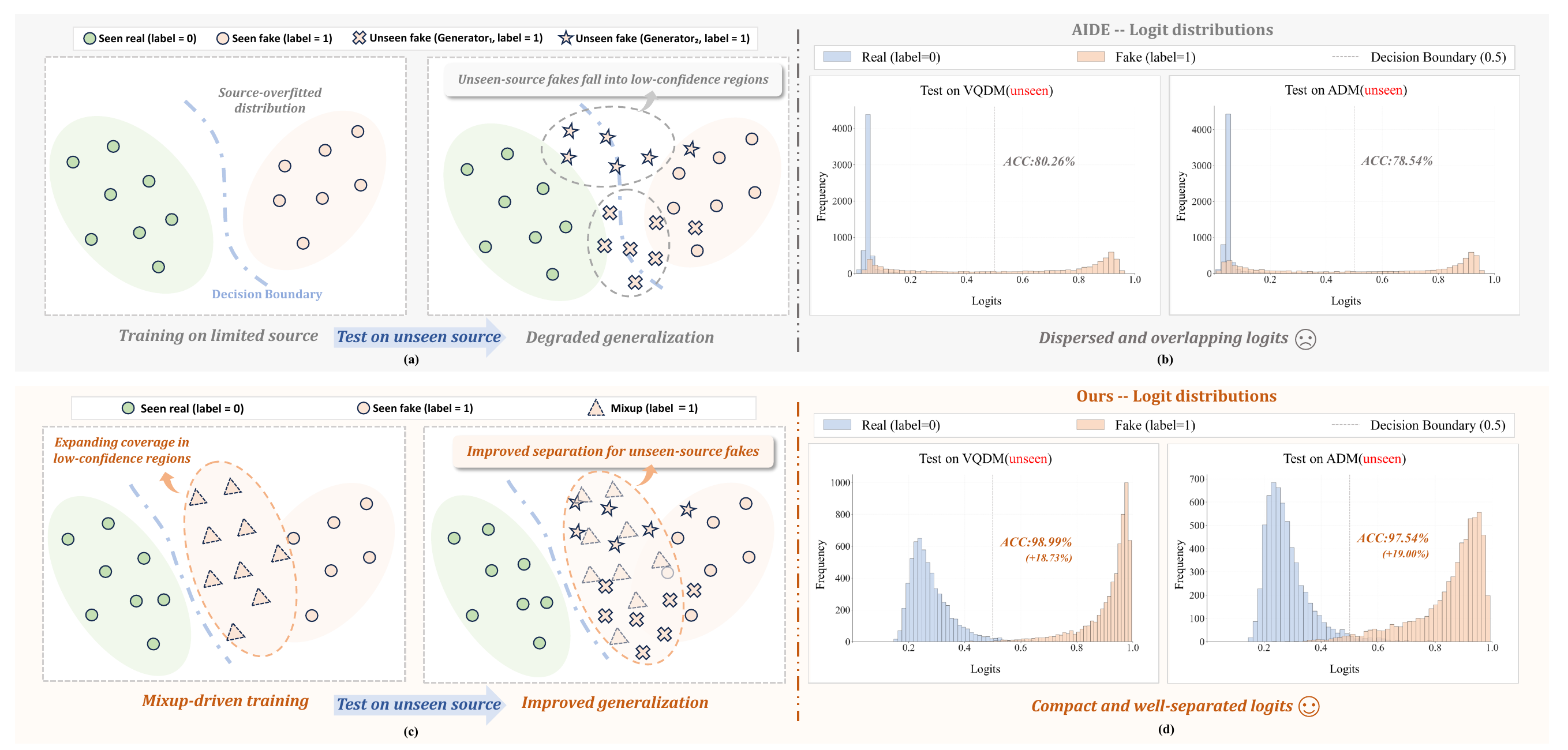}
  
  \caption{Comparison between traditional training and HiMix. 
(a) Training on limited sources yields clear separation within seen domains but induces source-overfitted distributions. As a result, fakes from unseen generators tend to fall into low-confidence regions (\textcolor{gray}{gray dashed ellipses}). 
(b) The logit distributions of AIDE~\cite{yansanity} on unseen VQDM~\cite{gu2022vector} / ADM~\cite{dhariwal2021diffusion} fakes exhibit less distinguishable predictions and dispersed logits, leading to degraded generalization. 
(c) In contrast, mixup-driven training constructs transitional samples that enrich low-confidence regions (\textcolor{orange}{orange dashed ellipses}), improving coverage of unseen-source fakes.
(d) HiMix produces more compact and well-separated logits, demonstrating improved generalization to unseen generators.}
  \label{fig:intro}
       \vspace{-7pt}
\end{figure*}

Early studies have explored image forgery detection using pixel-level features~\cite{zhong2023patchcraft,tan2024rethinking,wang2020cnn,rossler2019faceforensics++} or frequency-domain cues~\cite{frank2020leveraging,miao2023f,tan2024frequency,yansanity,chu2025fire}. Pixel-level methods exploit inter-pixel correlations~\cite{zhong2023patchcraft} or local upsampling artifacts~\cite{tan2024rethinking}, while frequency-based approaches capture subtle yet distinctive spectral differences between real and fake images. Although effective under matched training and testing sources, these methods rely on source-specific cues, leading to limited generalization. As illustrated in Figure~\ref{fig:intro}(a), training on limited sources induces source-overfitted distributions. Under distribution shift, unseen fake samples are often not well aligned with the learned fake distribution and lie in low-confidence regions, leading to degraded generalization. This phenomenon is further illustrated in Figure~\ref{fig:intro}(b), where logits produced by AIDE~\cite{yansanity} become more dispersed and less distinguishable on unseen domains, revealing the limitations of prior methods.

Building on this observation, recent SID studies~\cite{ojha2023towards,keita2025bi,liu2024forgery,cazenavette2024fakeinversion} have leveraged pretrained models to extract generic representations to improve detection performance on unseen generators. However, the extracted features mainly capture semantics tied to image content or category rather than intrinsic forgery cues~\cite{gye2025sfld,zhong2023patchcraft}, causing the detector to learn content-biased rather than generalizable patterns. Moreover, task-irrelevant components within these representations can obscure discriminative cues. To mitigate these issues, semantic perturbation strategies such as patch shuffling~\cite{gye2025sfld} and masking~\cite{li2025improving} have been introduced to weaken semantics and strengthen texture awareness, while task-oriented regularizations~\cite{zhang2025towards,chen2025forgelens} aim to suppress redundant information and emphasize forgery-relevant signals. Nevertheless, pretrained model-based approaches remain constrained at both the data and representation levels. From the data distribution perspective, training on limited generators leads to insufficient coverage of low-confidence regions. From the representation perspective, aggressive perturbations or regularization may disrupt consistent forgery patterns or introduce training-specific biases, ultimately compromising generalization.

To address these limitations, we propose a \textbf{H}ierarchical Art\textbf{i}fact-aware \textbf{Mix}up method for Synthetic Image Detection, termed \textit{HiMix}, which improves generalization from both data and representation perspectives. The \textbf{Mixup-driven Distributional Augmentation} (MDA) module constructs continuous transitional samples between real and fake images, effectively improving coverage of low-confidence regions where misclassification is most likely (Figure~\ref{fig:intro}(c)). By increasing the coverage of these regions and exposing the model to hard samples, MDA improves separability and generalization to unseen distributions. Meanwhile, the pixel-wise mixup operation achieves smooth semantic interpolation, reducing high-level semantic influence and highlighting fine-grained artifact cues for reliable detection. In parallel, the \textbf{Hierarchical Artifact-aware Representation} (HAR) module is designed to extract discriminative and robust forgery cues by aggregating artifact information from global and local perspectives. Through cross-layer integration and coarse-to-fine fusion, it captures complementary artifact patterns in complex mixed samples. As illustrated in Figure~\ref{fig:intro}(d), our method produces compact within-class (real/fake) and clearly separated between-class (real vs.\ fake) logit distributions, demonstrating enhanced generalization to unseen generators. The main contributions of this paper include:
\begin{itemize}
\item We propose HiMix, a unified detection framework composed of a Mixup-driven Distributional Augmentation (MDA) module and a Hierarchical Artifact-aware Representation (HAR) module, which jointly enhance generalization from both data and representation perspectives.
\item We introduce a distributional augmentation strategy that improves coverage of low-confidence regions while reducing the dominance of high-level semantics, coupled with a hierarchical fusion mechanism that captures complementary global and local artifact cues for robust representations.
\item Extensive experiments across multiple benchmarks demonstrate state-of-the-art performance, validating the cross-source generalization ability of our approach.
\end{itemize}

\section{Related Work}
\label{sec:related_work}

% \subsection{Pixel-based Detection}
% \label{subsec:image_based}
% Pixel-based detection methods directly operate in the spatial domain, aiming to identify forgery traces from local spatial patterns~\cite{zhong2023patchcraft,rossler2019faceforensics++,jia2025secret,liang2025transfer,nguyen2025forensic,liu2020global}.
% Early work by Wang et al.~\cite{wang2020cnn} showed that a standard CNN trained on a single generator, with proper data augmentation, can generalize effectively to unseen generators. Later, Tan et al.~\cite{tan2024rethinking} introduced NPR to capture structural inconsistencies caused by up-sampling operations. More recently, Li et al.~\cite{li2025improving} proposed the SAFE detector, approaching the problem from an image transformation perspective. Meanwhile, Jia et al.~\cite{jia2025secret} exploited the statistical uniformity of color distributions, integrating it with image features for improved synthetic image discrimination.

\subsection{Pixel-based Detection}
\label{subsec:image_based}
Pixel-based detection methods directly operate in the spatial domain, aiming to identify forgery traces from local spatial patterns~\cite{zhong2023patchcraft,rossler2019faceforensics++,jia2025secret,liang2025transfer,nguyen2025forensic,liu2020global}.
Early work by Wang et al.~\cite{wang2020cnn} showed that a standard CNN trained on a single generator, with proper data augmentation, can generalize effectively to unseen generators. Later, Tan et al.~\cite{tan2024rethinking} introduced NPR to capture structural inconsistencies caused by up-sampling operations. More recently, Li et al.~\cite{li2025improving} proposed the SAFE detector, approaching the problem from an image transformation perspective. Meanwhile, Jia et al.~\cite{jia2025secret} exploited the statistical uniformity of color distributions, integrating it with image features for improved synthetic image discrimination. Despite these advances, pixel-based methods often rely on generator-specific artifacts or local statistics, which may not generalize well under distribution shifts. 

\subsection{Frequency-based Detection}
\label{subsec:frequency_based}
Frequency-based detection methods exploit spectral artifacts, often from operations like up-sampling, by analyzing images in the frequency domain~\cite{frank2020leveraging,miao2023f,karageorgiou2025any,kashiani2025freqdebias,li2025optimized, wei2021universal}. Qian et al.~\cite{qian2020thinking} proposed F³-Net, a two-stream network that mines frequency cues via DCT-based image decomposition and local frequency statistics. Tan et al.~\cite{tan2024frequency} introduced FreqNet, which directly performs convolution on amplitude and phase spectra of high-frequency inputs. Meanwhile, FIRE~\cite{chu2025fire} and SPAI~\cite{karageorgiou2025any} approach the problem from a frequency reconstruction perspective, enhancing robustness against diverse forgery types. Recently, Yan et al.~\cite{yansanity} proposed AIDE, where the Patchwise Feature Extraction module leverages DCT-based scoring to select extreme-frequency patches for capturing subtle artifacts.

\subsection{Pre-trained Model-based Detection}
\label{subsec:pretrained_based}

Pre-trained model-based detection methods leverage models trained on web-scale datasets as powerful feature extractors~\cite{liu2024forgery,cazenavette2024fakeinversion,keita2025bi,yang2025d,zhou2025aigi,wei2024runge}. Ojha et al.~\cite{ojha2023towards} introduced UnivFD, which employs a frozen CLIP backbone to extract image features and performs classification via linear probing or nearest-neighbor search. To enhance CLIP’s representational capacity, Tan et al.~\cite{tan2025c2p} proposed C2P-CLIP, which injects category-related concepts through category common prompts and fine-tunes the image encoder using contrastive learning. From another perspective, Zhang et al.~\cite{zhang2025towards} developed VIB-Net, introducing a variational information bottleneck module after a frozen CLIP backbone to compress representations and retain only forgery-relevant information. Meanwhile, ForgeLens~\cite{chen2025forgelens} incorporates a weight-sharing guidance module to explicitly guide the extraction of forgery-specific features during training, mitigating the over-reliance on generic representations that often contain task-irrelevant information.

However, existing methods still struggle to generalize, as unseen forgeries often fall into low-confidence regions with less distinguishable predictions. HiMix enriches these regions via mixup and learns artifact-aware representations, yielding more stable and separable predictions.

\section{Method}
\label{sec:Method}

\begin{figure*}[]
  \centering
 
  {
        \includegraphics[width=1\linewidth]{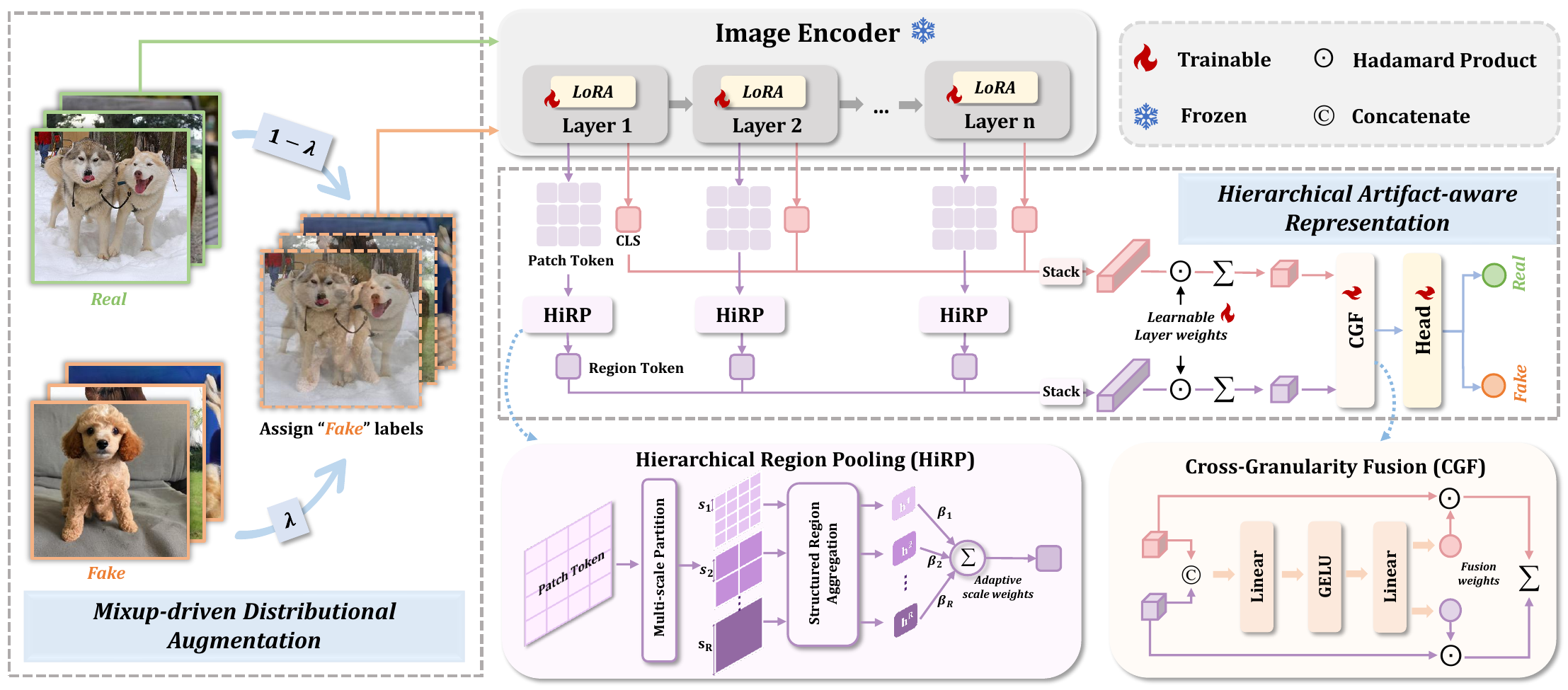}
     }
     
\caption{Overview of HiMix for generalized synthetic image detection. The framework integrates Mixup-driven Distributional Augmentation (MDA), which constructs pixel-level transitional samples between real and fake images to enrich low-confidence regions, with a Hierarchical Artifact-aware Representation (HAR) module. The latter generates region tokens via Hierarchical Region Pooling (HiRP), performs cross-layer fusion over CLS and region tokens, and further refines them through Cross-Granularity Fusion (CGF). With LoRA-based fine-tuning, HiMix achieves robust and generalized detection of unseen forgeries.}
       \vspace{-7pt}
   \label{fig:overview}
\end{figure*}

In this section, we will elaborate on the details of HiMix pipeline. The overall architecture is illustrated in \cref{fig:overview}. Our method enhances generalization through two collaborative modules: Mixup-driven Distributional Augmentation (MDA) and Hierarchical Artifact-aware Representation (HAR).

\subsection{Mixup-driven Distributional Augmentation}
To address the insufficient coverage of low-confidence regions caused by training on limited generative sources, we propose the MDA module. It constructs continuous transitional samples between real and synthetic images to expand the effective training distribution, particularly in low-confidence regions. Furthermore, the pixel-wise mixup operation performs smooth semantic interpolation, reducing high-level semantic influence while preserving essential structural cues, thereby encouraging the model to focus on low-level generator-invariant artifacts.

Specifically, inspired by Mixup~\cite{zhang2017mixup}, we randomly interpolate two samples $X_\text{real}$ and $X_\text{fake}$ drawn from real and fake images to generate a transitional sample $X_\text{mix}$, which is assigned to the fake class:
\begin{equation}
(X_{\text{mix}}, Y_{\text{mix}}) = (\lambda X_{\text{fake}} + (1-\lambda)X_{\text{real}},~ \textit{fake}),
\end{equation}
where $\lambda$ is sampled from a Beta distribution in $(0, 1)$. When $\lambda$ is small, the mixed sample is closer to the real distribution, while a larger $\lambda$ makes synthetic features more dominant. The resulting samples retain synthetic artifacts from $X_\text{fake}$ and are labeled as \textit{fake} for joint training with real images, providing diverse transitional and hard samples for robust learning.
\subsection{Hierarchical Artifact-aware Representation}
To learn discriminative artifact representations under distributional augmentation, we propose the HAR module. Built upon the CLIP image encoder, HAR employs hierarchical region pooling to extract multi-scale region tokens, and integrates them with the CLS token across layers to capture artifact cues at different depths. The resulting coarse- and fine-grained representations are further fused via a cross-granularity interaction mechanism, yielding a generator-invariant representation.

\noindent \textbf{Task-Adaptive Image Encoding}.
Benefiting from CLIP’s strong pretrained representations and efficient visual encoding capability, we adopt its image encoder as the backbone. Given an input image $X \in \mathbb{R}^{3 \times H \times W}$, the frozen CLIP~\cite{radford2021learning} divides it into a series of patches and processes them through Transformer layers. The output $\mathbf{Z}_l \in \mathbb{R}^{(N+1) \times d}$ of the $l$-th layer is formulated as:
\begin{equation}
\mathbf{Z}_l = \left[ \mathbf{z}_l^\mathrm{cls}, \mathbf{z}_l^{p_1}, \mathbf{z}_l^{p_2}, \ldots, \mathbf{z}_l^{p_N} \right],
\end{equation}
where $\mathbf{z}_l^\mathrm{cls}$ denotes the CLS token capturing the global representation, and $\mathbf{z}_l^{p_i}$ represents the patch token encoding local texture information.

As pretrained CLIP primarily captures content-related semantic information, which may overshadow subtle forgery cues, we introduce a lightweight task-adaptive fine-tuning strategy to mitigate such semantic bias. Specifically, we integrate LoRA~\cite{hu2022lora} into the attention mechanism, performing low-rank adaptation on the QKV projections as follows:

\begin{equation}
\left\{\mathbf{Q'},\mathbf{K'},\mathbf{V'}\right\} = \left\{\mathbf{Q},\mathbf{K},\mathbf{V}\right\} + \mathbf{A}_{QKV}\mathbf{B}_{QKV},
\end{equation}
where $\mathbf{A}_{QKV}$ and $\mathbf{B}_{QKV}$ are learnable low-rank adaptation matrices applied to the QKV projections. 

Through this lightweight adaptation, the model adjusts its attention toward forgery-relevant cues while preserving the integrity of pretrained representations, enabling robust multi-level and fine-grained artifact learning.

\noindent \textbf{Hierarchical Region Pooling.}
To effectively capture the structural regularities of forgery artifacts, we introduce the Hierarchical Region Pooling (HiRP) module. At its core, HiRP aggregates local statistics across multiple spatial scales by partitioning the patch-token grid into local windows, enhancing the representation of subtle and spatially consistent forgery cues.

Given the patch tokens $\mathbf{Z}_l^{p} \in \mathbb{R}^{N \times d}$ from the $l$-th layer, we reshape them into a 2D grid of size $h \times w$, where $N = hw$. We then perform fine-grained multi-scale spatial partition using $R$ window sizes $\{s_r\}_{r=1}^R$. At each scale $r$, the grid is uniformly divided into $M_r = \frac{hw}{s_r^2}$ non-overlapping local windows, denoted as $\{\mathbf{p}^{r}_i\}_{i=1}^{M_r}$, where $\mathbf{p}^{r}_i \in \mathbb{R}^{s_r \times s_r \times d}$ represents the local representations within the $i$-th window.

To effectively summarize region-level statistics, we introduce a lightweight Structured Region Aggregation operation, which consists of two sequential steps: intra-window average pooling ($\operatorname{AvgPool}$) and inter-window max pooling ($\operatorname{MaxPool}$). Specifically, we first aggregate each local window $\mathbf{p}^{r}_i$ via $\operatorname{AvgPool}$ to obtain stable local representations, followed by $\operatorname{MaxPool}$ across all $M_r$ windows to capture the most salient responses, yielding the scale-level representation $\mathbf{h}_l^{r} \in \mathbb{R}^d$:
\begin{equation}
\mathbf{h}_l^{r} = \operatorname{MaxPool} \Big( \big\{ \operatorname{AvgPool}(\mathbf{p}^{r}_i) \big\}_{i=1}^{M_r} \Big).
\end{equation}
% This intra-average and inter-max strategy ensures that representations extracted across different scales remain highly discriminative.

Finally, the scale-level representations are adaptively fused into a comprehensive regional token $\mathbf{z}_l^{\mathrm{reg}} \in \mathbb{R}^d$:
\begin{equation}
\mathbf{z}_l^{\mathrm{reg}} = \sum_{r=1}^{R} \beta_r \mathbf{h}_l^{r}, \quad \sum_{r=1}^{R} \beta_r = 1, 
\end{equation}
where $\beta_r$ are learnable weights, normalized via softmax, that dynamically control the importance of each spatial scale.

\noindent \textbf{Cross-Layer Fusion.}  
As forgery cues emerge at different semantic depths, cross-layer fusion enables the joint aggregation of fine-grained artifacts and high-level structural consistency.

Given the CLS token $\mathbf{z}_l^{\mathrm{cls}}$ and region token $\mathbf{z}_l^{\mathrm{reg}}$ extracted from the $l$-th selected layer, we first stack the tokens of the same type across layers to form a layer-wise token matrix:
\begin{equation}
\mathbf{Z}^t = \operatorname{Stack}(\mathbf{z}_1^t,\mathbf{z}_2^t,\ldots,\mathbf{z}_n^t)\in\mathbb{R}^{n\times d},
\end{equation}
where $t\in\{\mathrm{cls},\mathrm{reg}\}$ and $n$ denotes the number of selected layers.

To account for the varying importance of different layers, we introduce a learnable weight vector $\mathbf{a}^t=[a_1^t,a_2^t,\ldots,a_n^t]^\top$, which is normalized by
\begin{equation}
\hat{\mathbf{a}}^t=\operatorname{Softmax}(\mathbf{a}^t).
\end{equation}

The fused representation is then obtained by a weighted summation over the layer-wise features:
\begin{equation}
\tilde{\mathbf{z}}^t=(\hat{\mathbf{a}}^t)^\top \mathbf{Z}^t.
\end{equation}

\noindent \textbf{Cross-Granularity Fusion.}
To jointly integrate global structural patterns and local artifact cues, we introduce a lightweight Cross-Granularity Fusion (CGF) module. 
Given the global feature $\tilde{\mathbf{z}}^\mathrm{cls}$ and the local feature $\tilde{\mathbf{z}}^\mathrm{reg}$, we first concatenate them and feed the result into a multi-layer perceptron (MLP) to predict adaptive fusion weights:
\begin{align}
\mathbf{w} &= \operatorname{Softmax}\big(\operatorname{MLP}([\tilde{\mathbf{z}}^\mathrm{cls}, \tilde{\mathbf{z}}^{reg}])\big),
\end{align}
where $[\cdot,\cdot]$ denotes concatenation and $\mathbf{w}=(w_\mathrm{cls},w_\mathrm{reg})\in\mathbb{R}^{2}$. 
The MLP consists of two linear layers with a GELU activation in between.
The final fused representation is computed as:
\begin{align}
\tilde{\mathbf{z}}^{f} &= w_\mathrm{cls}\tilde{\mathbf{z}}^{cls} + w_\mathrm{reg}\tilde{\mathbf{z}}^\mathrm{reg}.
\end{align}

\begin{table*}[t]
\caption{Evaluation on GenImage dataset. Comparison of accuracy (Acc) / average precision (AP) (\%) with state-of-the-art methods on the GenImage \citep{zhu2023genimage} dataset.}
\small \centering
    \renewcommand{\arraystretch}{1.2}
    \scalebox{0.96}[0.96]{
\begin{tabular}{lcccccccccc}
\toprule
\textbf{Method} &{\textbf{Ref}} &{\textbf{Midjourney}} &\textbf{{SD v1.4}} & \textbf{{SD v1.5}} & \textbf{{ADM}} &\textbf{{GLIDE}} &\textbf{{Wukong}} &\textbf{{VQDM} }&\textbf{{BigGAN}} & {\textbf{\textit{Mean}}} \\ \midrule

{CNNDect} &CVPR 20\citep{wang2020cnn} & 50.1 / 53.4 & 50.3 / 55.9 & 50.3 / 56.1 & 53.0 / 69.2  & 51.7 / 66.9 & 51.4 / 62.4  & 50.0 / 53.5 & 69.8 / 91.5  & 53.3 / 63.6 \\
{LGrad} &CVPR 23\citep{tan2023learning}  & 73.7 / 77.5 & 76.3 / 80.1 & 77.4 / 80.1 & 51.8 / 51.0  & 49.8 / 50.5 & 73.1 / 75.4  & 52.1 / 51.5 & 40.5 / 30.2  & 61.8 / 62.0 \\
{UniFD}  &CVPR 23\citep{ojha2023towards} & 56.9 / 68.5 & 65.1 / 81.5 & 64.7 / 81.0 & 69.2 / 84.2  & 60.1 / 73.5 & 73.5 / 89.0  & 86.0 / 95.0 & 89.3 / 97.0  & 70.6 / 83.7 \\
{NPR}  &CVPR 24\citep{tan2024rethinking}  & 77.8 / 85.4 & 78.6 / 84.0 & 78.9 / 84.6 & 69.7 / 74.6  & 78.4 / 85.7 & 76.1 / 80.5  & 78.1 / 81.0 & 80.1 / 88.2  & 77.2 / 83.0 \\
{FreqNet} &AAAI 24\citep{tan2024frequency} & 69.7 / 78.5 & 64.2 / 74.5 & 64.9 / 75.6 & 83.5 / 92.0  & 81.2 / 88.5 & 57.8 / 67.0  & 81.4 / 90.0 & 90.5 / 95.0  & 74.1 / 82.6 \\
{AIDE} &ICLR 25\citep{yansanity} & 81.4 / 98.0 & 99.8 / 100.0 & 99.8 / 100.0 & 78.5 / 94.6  & 91.8 / 99.1 & 98.9 / 100.0  & 80.2 / 97.1 & 66.8 / 93.5  & 87.2 / 97.8 \\
{SAFE} &KDD 25\citep{li2025improving} & 95.2 / 99.0 & 99.4 / 99.1 & 99.3 / 99.7 & 82.2 / 96.7  & 96.2 / 99.3 & 98.1 / 99.8  & 96.2 / 99.4 & 97.7 / 99.8  & 95.5 / 99.1 \\
{VIB} &CVPR 25\citep{zhang2025towards}  & 88.3 / 98.4  & 99.5 / 100.0 & 99.3 / 100.0  & 74.3 / 91.2  & 73.4 / 95.1  & 98.7 / 99.9  & 89.4 / 97.3 & 55.6 / 85.4 & 84.8 / 95.9 \\
{CoD} &CVPR 25\citep{jia2025secret}  & 96.0 / 99.0 & 99.7 / 99.9 & 99.8 / 100.0 & 85.2 / 97.4 & 95.9 / 99.2 & 98.2 / 99.9 & 96.8 / 99.9 & 98.3 / 99.9 & 96.2 / 99.4 \\ 
\rowcolor{lightblue}
\midrule
\textit{\textbf{HiMix}}& \textit{Ours} &91.9 / 98.8 & 99.0 / 100.0 & 99.0 / 100.0 & 97.5 / 99.7  & 99.1 / 99.9 & 99.1 / 100.0  & 99.0 / 99.9& 95.9 / 99.4  & \textbf{97.6 / 99.7} \\ 
\bottomrule
\end{tabular}
}

\label{table:b1}
% \vspace{-6mm}
\end{table*}

\subsection{Prediction and Optimization}
After obtaining the final fused representation $\tilde{\mathbf{z}}^{f}$, we employ a lightweight classification head to predict the final score:
\begin{equation}
\hat{y}_i = \operatorname{Head}\left(\tilde{\mathbf{z}}^{f}\right),
\end{equation}
where $\operatorname{Head}(\cdot)$ denotes a simple classifier composed of two ReLU-activated linear layers followed by a sigmoid output.

The network is optimized using the binary cross-entropy loss:
\begin{equation}
\mathcal{L}_{ce} = -\frac{1}{b}\sum_{i=1}^{b} \left[y_i \log \hat{y}_i + (1 - y_i)\log (1 - \hat{y}_i)\right],
\end{equation}
where $b$ denotes the batch size, and $y_i \in \{0, 1\}$ is the ground-truth label, with 0 indicating a real image and 1 indicating a fake image.

\section{Experiments}\label{Experiments}
\subsection{Experimental Setup}

\noindent \textbf{Training Datasets.} Following GenImage~\cite{zhu2023genimage}, we use 162,000 ImageNet~\cite{russakovsky2015imagenet} real images and 162,000 Stable Diffusion v1.4~\cite{rombach2022high} images. The generation of this synthetic set is conditioned on the 1,000 ImageNet class labels, yielding 162 images per label. Following ForenSynths~\cite{wang2020cnn}, we use 72,000 LSUN~\cite{yu2015lsun} real images and 72,000 ProGAN~\cite{karras2017progressive} images.This generation process is based on 4 distinct LSUN labels, resulting in 18,000 images for each class.

% Our implementation is based on the code of CoOp~\cite{zhou2022learning}. 
\noindent \textbf{Evaluation Datasets.} We evaluate our method on four representative benchmarks: GenImage~\cite{zhu2023genimage}, a large-scale dataset containing real and synthetic images from eight diverse generators. Ojha~\cite{ojha2023towards}, featuring diffusion and autoregressive-based forgeries from ImageNet and LAION. ForenSynths~\cite{wang2020cnn}, designed to test cross-generator generalization across multiple CNN-based models including deepfakes. TwinSynths-GAN~\cite{gye2025sfld}, a high fidelity benchmark that pairs each real image with a content consistent synthetic counterpart for fair and challenging evaluation. \textit{For more details, please refer to the supplementary material.}

\noindent \textbf{Evaluation Metric.}  
To ensure a fair and direct comparison with prior studies~\cite{li2025improving,jia2025secret}, we follow the same evaluation protocol. Specifically, the performance of our model is measured using two widely adopted metrics: accuracy (Acc) and average precision (AP), with the decision threshold set to 0.5.

\noindent \textbf{Implementation Details.}  
We employ the Adam optimizer~\cite{kingma2014adam} with an initial learning rate of $1\times 10^{-5}$ and a batch size of 256. The image encoder is initialized from the CLIP ViT-L/14 model~\cite{radford2021learning} as a feature extractor. LoRA layers are applied to the projection matrices $\mathbf{Q}_{\text{proj}}$, $\mathbf{K}_{\text{proj}}$, and $\mathbf{V}_{\text{proj}}$, with hyperparameters set to $r_\text{LoRA}=8$ and $\alpha_\text{LoRA}=16$, enabling task-adaptive fine-tuning with minimal additional parameters. For mixup-based augmentation, the interpolation ratio is sampled as $\lambda \sim \text{Beta}(\alpha, \alpha)$ with $\alpha=0.1$~\cite{zhang2017mixup}, which achieves the best performance in our ablation studies. For hierarchical region pooling, we use $R=3$ scales with window sizes $s_r \in \{2,4,8\}$. During training, input images are randomly cropped to $224\times224$ and horizontally flipped with a probability of 0.5. At test time, only center cropping with resolution $224\times224$ is applied. All experiments are implemented in PyTorch~\cite{paszke2019pytorch} and conducted on four NVIDIA RTX 3090 GPUs.

\begin{table*}[t]
\caption{Evaluation on diffusion model dataset. Comparison of accuracy (Acc) / average precision (AP) (\%) with state-of-the-art methods on the diffusion models from Ojha \citep{ojha2023towards} dataset.}
% \vspace{-1mm}
\small \centering
    \renewcommand{\arraystretch}{1.21}
    \scalebox{0.91}[0.91]{
\begin{tabular}{lcccccccccc}
\toprule
\textbf{Method} &{\textbf{Ref}} &{\textbf{DALLE}} &\textbf{Glide\_100\_10} & \textbf{{Glide\_100\_27}} & \textbf{{Glide\_50\_27}} &\textbf{{ADM}} &\textbf{{LDM\_100}} &\textbf{{LDM\_200}}&\textbf{{LDM\_200\_cfg}} & {\textbf{\textit{Mean}}} \\ \midrule

{CNNDect} &CVPR 20\citep{wang2020cnn} & 51.8 / 61.2 & 53.3 / 72.9 & 53.1 / 71.3 & 54.2 / 76.1  & 54.9 / 66.6 & 51.9 / 63.7  & 52.0 / 64.5 & 51.6 / 63.1  & 52.8 / 67.5 \\
{LGrad} &CVPR 23\citep{tan2023learning}  & 88.5 / 97.3 & 89.5 / 94.9 & 87.4 / 93.2 & 90.6 / 95.0  & 86.7 / 99.8 & 94.8 / 99.2  & 94.2 / 99.1& 95.8 / 99.2  & 90.9 / 97.3 \\
{UniFD}  &CVPR 23\citep{ojha2023towards} & 89.3 / 96.5 & 90.0 / 96.5 & 90.7 / 97.0 & 90.8 / 97.2  & 75.1 / 84.5 & 90.1 / 97.0  & 90.2 / 97.0& 77.3 / 88.6  & 86.7 / 94.3 \\
{NPR}  &CVPR 24\citep{tan2024rethinking}  & 94.5 / 99.5 & 98.2 / 99.8 & 97.8 / 99.7 & 98.2 / 99.8  & 75.8 / 81.0 & 99.3 / 99.0  & 99.1 / 99.9& 99.0 / 99.9  & 95.1 / 97.4 \\
{FreqNet} &AAAI 24\citep{tan2024frequency} & 97.2 / 99.5 & 87.8 / 96.1 & 84.4 / 96.6 & 86.5 / 95.0  & 67.2 / 74.5 & 97.8 / 99.6  & 97.4 / 99.0& 98.2 / 99.0  & 89.6 / 94.9 \\
{SAFE} &KDD 25\citep{li2025improving} & 97.5 / 99.7 & 97.2 / 99.4 & 95.8 / 98.9 & 96.6 / 99.2  & 82.4 / 95.7 & 98.8 / 100.0  & 98.8 / 100.0& 98.7 / 99.8  & 95.7 / 99.0 \\
{C2P-CLIP} &AAAI 25\citep{tan2025c2p} &98.6 / 99.9 	&96.1 / 99.8 	&95.3 / 99.7	&95.3 / 99.8	&69.1 / 94.1	&99.3 / 100.0	&99.3 / 100.0	&97.3 / 99.8	&93.8 / 99.1 \\
{VIB} &CVPR 25\citep{zhang2025towards}  & 87.8 / 96.8 & 86.9 / 97.6 & 86.2 / 97.5 & 89.5 / 98.1  & 71.4 / 86.7 & 96.5 / 99.4 & 96.2 / 99.4 & 77.7 / 92.9 & 86.5 / 96.1 \\
{CoD} &CVPR 25\citep{jia2025secret}  & 98.1 / 99.6 & 97.6 / 99.6 & 97.4 / 99.5 & 97.7 / 99.5 & 92.8 / 97.4 & 98.8 / 99.8 & 98.9 / 100.0& 98.7 / 99.8 & 97.5 / 99.4 \\ 
\rowcolor{lightblue}
\midrule
\textit{\textbf{HiMix}}& \textit{Ours} &98.5 / 99.9 & 98.5 / 99.9 & 98.5 / 99.9 & 98.4 / 99.8  & 97.5 / 99.7 & 98.5 / 100.0  & 98.5 / 100.0& 98.4 / 100.0  & \textbf{98.3 / 99.9} \\ 
\bottomrule
\end{tabular}
}

\label{table:b2}
% \vspace{-6mm}
\end{table*}

\begin{table}[]
\vspace{-3pt}
\caption{Evaluation on TwinSynths-GAN dataset. Comparison of average precision (AP) (\%) with state-of-the-art methods on TwinSynths-GAN \citep{gye2025sfld} dataset.}
\small \centering
\setlength{\tabcolsep}{12pt} % 控制列间距（默认6pt，建议8~10pt）
    \scalebox{0.95}[0.95]{
\begin{tabular}{lcc}
\toprule
\textbf{Method} &{\textbf{Ref}} &{\textbf{TwinSynths-GAN}} \\ \midrule

{CNNDect} &CVPR 20\citep{wang2020cnn} & 62.9  \\
{FreDect} &CVPR 20\citep{frank2020leveraging} & 54.6  \\
{GramNet} &CVPR 20\citep{liu2020global} & 72.0  \\
{Fusing} &CVPR 22\citep{ju2022fusing} & 61.8  \\
{LGrad} &CVPR 23\citep{tan2023learning} & 59.5  \\
{UniFD}  &CVPR 23\citep{ojha2023towards} & 58.1  \\
{NPR}  &CVPR 24\citep{tan2024rethinking} & 78.2  \\
{AIDE} &ICLR 25\citep{yansanity} & 52.4  \\
{SAFE} &KDD 25\citep{li2025improving} & 46.6  \\
{VIB} &CVPR 25\citep{zhang2025towards} & 52.1  \\
{SFLD} &CVPR 25\citep{gye2025sfld} & 73.8  \\
\rowcolor{lightblue}
\midrule
\textit{\textbf{HiMix}}& \textit{Ours} &\textbf{90.6} \\ 
\bottomrule
\end{tabular}
}

\label{table:b4}
    \vspace{-3pt}
\end{table}

\subsection{Main Results}
\noindent \textbf{Evaluation on GenImage Dataset.}    
We first evaluate HiMix on the GenImage dataset, which contains forged images synthesized by eight GAN and diffusion models. As shown in \cref{table:b1}, HiMix achieves an average Acc of 97.6\% and an average AP of 99.7\%, outperforming strong baselines such as SAFE~\cite{li2025improving} and CoD~\cite{jia2025secret}. Notably, while existing detectors often struggle with diffusion-based forgeries such as ADM due to refined noise scheduling and high photorealism, HiMix maintains consistently strong performance across different generators. This robustness stems from the mixup-driven augmentation, which better covers low-confidence regions and exposes the model to diverse transitional and hard samples, and the hierarchical artifact-aware representation, which captures complementary forgery cues across multiple granularities.

% \noindent \textbf{Evaluation on GenImage Dataset.} We first evaluate HiMix on the GenImage Dataset, which contains forged images synthesized by eight GAN and diffusion models. As shown in \cref{table:b1}, HiMix achieves an average Acc of 97.6\% and an average AP of 99.7\%, outperforming strong baselines such as SAFE~\cite{li2025improving} and CoD~\cite{jia2025secret}. While existing detectors often struggle with diffusion-based forgeries like ADM due to refined noise scheduling and photorealistic quality, HiMix remains robust, benefiting from mixup-driven augmentation and hierarchical artifact fusion to capture subtle artifacts and maintain stable decision boundaries across generators.

\noindent \textbf{Evaluation on Diffusion Model Dataset.} We further evaluate HiMix on a DM-based dataset synthesized by multiple diffusion architectures. Trained solely on SDv1.4, HiMix generalizes effectively within the diffusion family such as LDM and Glide, achieving an average Acc of 98.3\% and an average AP of 99.9\% as shown in \cref{table:b2}. These results indicate that HiMix not only maintains high accuracy across diffusion variants but also exhibits strong resilience to generator-specific biases. This demonstrates that even within the same diffusion family, our approach avoids overfitting to the training generator and effectively captures generator-invariant forgery cues, ensuring robust and transferable detection across diverse diffusion variants.

% \noindent \textbf{Evaluation on GAN-based dataset.} 
% Next, we evaluate HiMix on the GAN-based dataset.  As shown in \cref{table:b3}, our method achieves an Acc of 96.9\% and an AP of 99.7\%, ranking second in accuracy but reaching the highest AP among all competitors. This indicates that HiMix produces more reliable confidence calibration, capturing subtle distinctions between real and fake samples even when decision boundaries overlap. Notably, on the DeepFake subset, our model achieves the best detection accuracy, verifying the effectiveness of the hierarchical fusion module through cross-layer integration and coarse-to-fine granularity fusion. Overall, even when trained solely on ProGAN, the model demonstrates strong cross-generator generalization and robustness to semantically complex real-world forgeries.
\noindent \textbf{Evaluation on GANs Dataset.} 
 To further assess generalization, we evaluate our model on TwinSynths-GAN, which produces high-fidelity, content-aligned forgeries that are nearly indistinguishable from real images. In this challenging setting, most detectors suffer severe degradation, while HiMix attains an average AP of 90.48\%, exceeding the latest SOTA methods by 12.29\% as shown in \cref{table:b4}. This improvement stems from MDA exposing the model to hard samples and HAR integrating multi-scale cues to capture subtle artifacts. These results show that HiMix maintains reliable detection ability, even for highly realistic forgeries. For the traditional GAN-based ForenSynths~\cite{wang2020cnn} dataset, HiMix maintains performance comparable to SOTA detectors, \textit{as detailed in the supplementary material}.

% \subsection{Analysis and Discussion}
% \label{sec:analysis}

% \begin{table}[h]
% \caption{Low-FPR operating-point behavior. RFPR denotes Real FPR. All @1\% metrics use a single GenImage-calibrated threshold (1\% FPR). RFPR and TPR are reported in \%, ECE is unitless.}
% \centering
% \footnotesize
% \setlength{\tabcolsep}{5pt}
% \renewcommand{\arraystretch}{1.15} % 控制行距（默认1.0，1.2~1.3效果更好）
% \scalebox{0.95}[0.95]{
% \begin{tabular}{l c c c c}
% \toprule
% \multirow{2}{*}{\textbf{Method}} 
% & \textbf{RFPR @ 1\%} $\downarrow$
% & \textbf{RFPR @ 1\%} $\downarrow$
% & \textbf{TPR @ 1\%} $\uparrow$
% & \textbf{ECE} $\downarrow$ \\
% & \textbf{WildRF}
% & \textbf{FaceForensics++}
% & \textbf{GenImage}
% & \textbf{GenImage} \\
% \midrule
% AIDE & 9.45 & 14.48 & 84.27 & 0.12 \\
% SAFE & 0.81 & 1.85 & 93.10 & 0.07 \\
% \textbf{Ours} & \textbf{0.75} & \textbf{1.13} & \textbf{94.76} & \textbf{0.05} \\
% \bottomrule
% \end{tabular}}
% \label{tab:low_fpr}
% \end{table}

\subsection{Analysis and Discussion}
\label{sec:analysis}

\noindent \textbf{Analysis of Mixup-Driven Augmentation.}
\label{sec:mda_analysis}
We further analyze the Mixup-driven Distributional Augmentation (MDA) module to address two potential concerns: \textit{(1) whether gains stem from mixup-specific artifacts rather than genuine synthetic traces, and (2) whether assigning hard ``fake'' labels to near-real mixtures inflates false positives on authentic images.}

\noindent \textbf{(1) Robustness to Mixup-Induced Artifacts.} 
Linear interpolation may introduce mixup-specific artifacts, risking shortcut learning. We first apply \textit{real-real mixup} as a control (\cref{tab:robustness_analysis}, top). If such artifacts were exploited, labeling mixed real images as ``fake'' would impair real-image recognition. Instead, the Real FPR on FaceForensics++ (FF++) increases only marginally (from 1.13\% to 1.41\%), indicating minimal reliance on mixup artifacts. To rule out reliance on simple spatial perturbation–based augmentations, we test \textit{patch shuffling}. While it avoids interpolation artifacts by simply disrupting spatial structures, its performance is significantly inferior (GenImage TPR@1\% drops to 85.69\%, RFPR doubles). This confirms HiMix effectively captures intrinsic generative traces rather than exploiting mixup-specific or generic augmentation artifacts.

\begin{table}[h]
\vspace{-3pt}
\caption{Analysis of augmentation robustness and low-FPR behavior.
\textit{Top:} Controls for interpolation artifacts using alternative augmentations. \textit{Bottom:} Low-FPR behavior and calibration under a fixed operating point. RFPR denotes Real FPR. RFPR and TPR are reported in  \%, and ECE is unitless. All @1 \% metrics use a single threshold calibrated on GenImage. WRF: WildRF, FF++: FaceForensics++.}
\small
\centering
% \footnotesize
\setlength{\tabcolsep}{5pt} % 表头缩写后，可以适当增加列间距让表格更舒展
\renewcommand{\arraystretch}{1.05}
\scalebox{0.84}[0.84]{
\begin{tabular}{l c c c c}
\toprule
\multirow{2}{*}{\textbf{Method}} 
& \textbf{RFPR @ 1\%} $\downarrow$
& \textbf{RFPR @ 1\%} $\downarrow$
& \textbf{TPR @ 1\%} $\uparrow$
& \textbf{ECE} $\downarrow$ \\
& \textbf{WRF}
& \textbf{FF++}
& \textbf{GenImage}
& \textbf{GenImage} \\
\midrule
\multicolumn{5}{c}{\textit{Controls for Interpolation Artifacts}} \\
\midrule
Real-Real Mixup & 1.07 & 1.41 & 79.56 & 0.09 \\
Patch Shuffling & 3.68 & 2.60 & 85.69 & 0.15 \\
\midrule
\multicolumn{5}{c}{\textit{Low-FPR Behavior and Calibration}} \\
\midrule
AIDE & 9.45 & 14.48 & 84.27 & 0.12 \\
SAFE & 0.81 & 1.85 & 93.10 & 0.07 \\
\midrule
\textbf{\textit{HiMix} (Ours)} & \textbf{0.75} & \textbf{1.13} & \textbf{94.76} & \textbf{0.05} \\
\bottomrule
\end{tabular}}

\label{tab:robustness_analysis}
\vspace{-11pt}
\end{table}

\noindent \textbf{(2) Mitigating False Positives from Hard Labeling.} 
Assigning a hard ``fake'' label to near-real mixtures (i.e., mixtures with small $\lambda$) raises the concern of biasing the classifier toward the fake class, thereby inflating false positives on authentic images. We dispel this concern by evaluating the model's strict low-FPR behavior and calibration (\cref{tab:robustness_analysis}, bottom). Using a single threshold calibrated at 1\% FPR on GenImage, HiMix achieves exceptionally low RFPR@1\% on unseen real-only datasets (WRF: 0.75\%, FF++: 1.13\%), significantly outperforming baselines like AIDE and SAFE. Furthermore, it yields a highly stable Expected Calibration Error (ECE) of just 0.05 on GenImage. These robust metrics demonstrate that our labeling strategy effectively enriches low-confidence regions without biasing the model against the real image manifold.

\begin{figure*}[h]
  \centering
  {
        \includegraphics[width=0.95\linewidth]{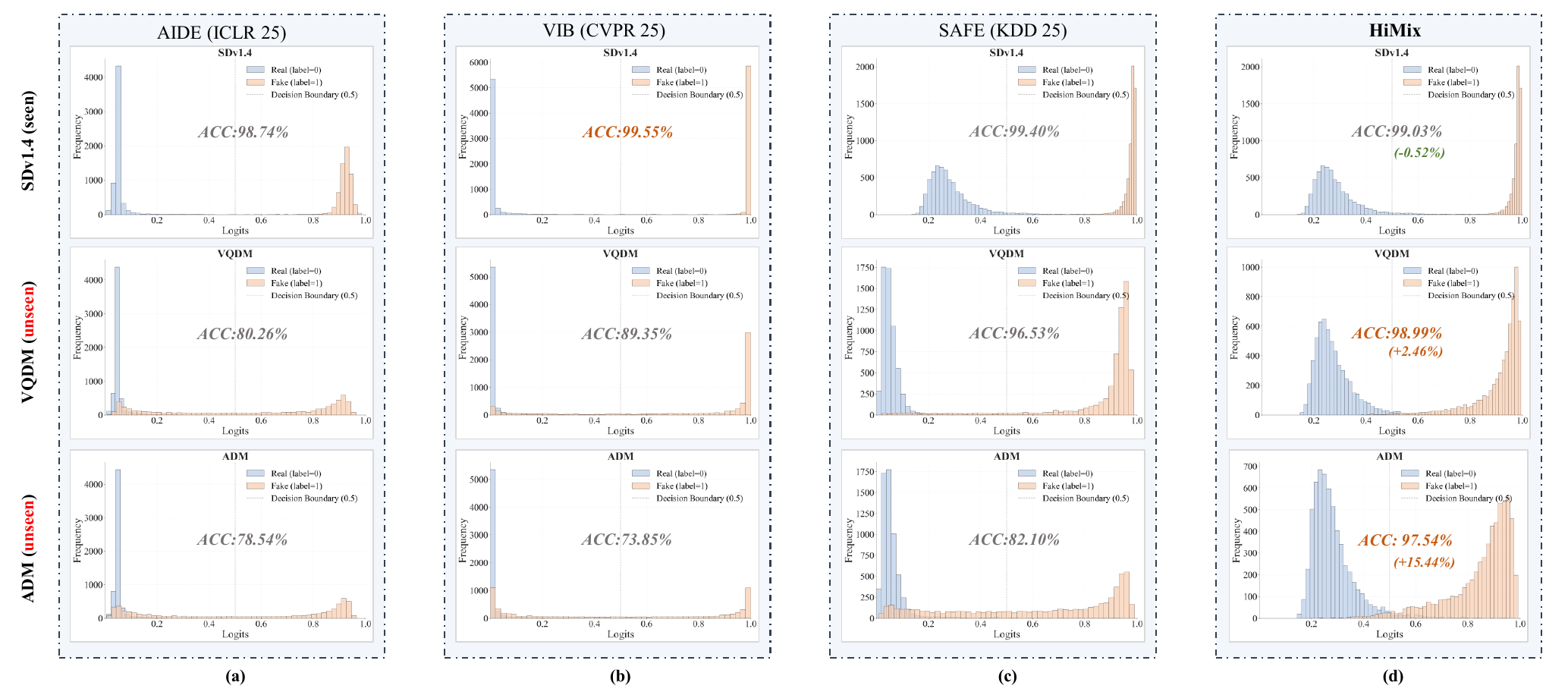}
     
     }
   \caption{Logit distributions of real and fake images on SDv1.4 (seen generator) and VQDM, ADM (unseen generators). With the decision threshold set to 0.5, HiMix produces compact within-class and clearly separated between-class logit distributions compared with state-of-the-art detectors AIDE~\cite{yansanity}, VIB~\cite{zhang2025towards}, and SAFE~\cite{li2025improving}, demonstrating superior cross-source generalization.}
 
   \label{fig:logit}
   % \vspace{-10pt}
\end{figure*}

\subsection{Visualization} 
\noindent \textbf{Logit Distribution Analysis.} To further evaluate the generalization capability, we visualize the logit distributions of our method and state-of-the-art approaches, as shown in Figure~\ref{fig:logit}. Existing methods perform well on seen datasets, however, their logits become scattered on unseen generators, spreading across the full probability range [0,1] with highly unstable decision boundaries and noticeable overlap between real and fake regions, leading to frequent misclassification of forgeries as real images. In contrast, our method maintains compact within-class (real/fake) and clearly separated between-class (real vs.\ fake) logit distributions, even on unseen sources, demonstrating consistently high accuracy and robust generalization.

\begin{figure}[t]
  \centering
  {
        \includegraphics[width=1\linewidth]{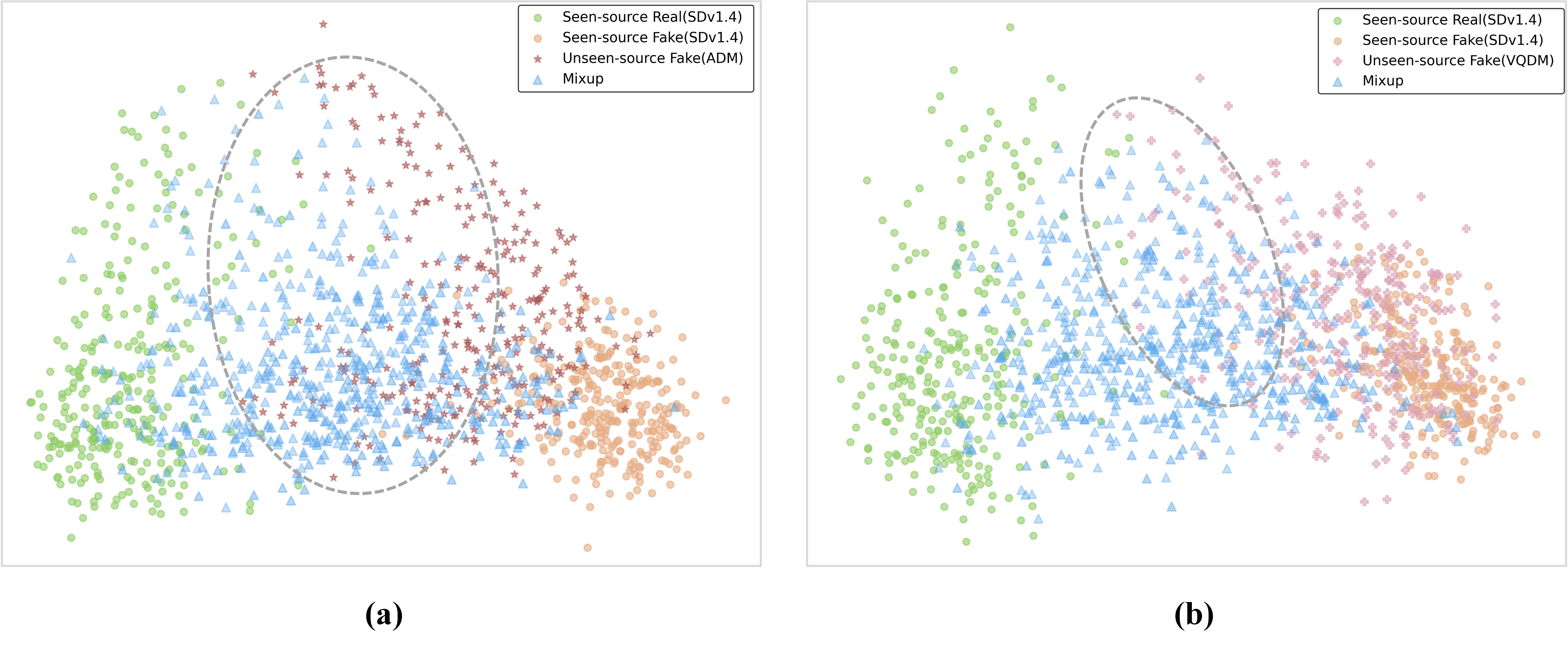}
     
     }
\caption{t-SNE visualization of MDA. Seen real (\textcolor{greennew}{green circles}) and fake (\textcolor{orangenew}{orange circles}) samples are well separated. Unseen forgeries from ADM (\textcolor{brownnew}{reddish-brown pentagrams}) and VQDM (\textcolor{pink}{pink crosses}) lie in low-confidence regions (\textcolor{gray}{gray dashed ellipses}). Mixup samples (\textcolor{bluenew}{blue triangles}) form transitional samples that enrich these regions, improving coverage of unseen forgeries and leading to more stable and separable distributions.}

 \vspace{-5pt}
   \label{fig:tsne}
   % \vspace{-10pt}
\end{figure}

\noindent \textbf{Mixup-driven Visualization.}  
To better understand the effect of our mixup-driven augmentation, we visualize the feature distributions in a 2D embedding space (Figure~\ref{fig:tsne}). Seen-source real and fake samples generated by SDv1.4 are shown as green and orange circles, mixup samples produced by the MDA module as blue triangles, and unseen-source fakes from ADM (reddish-brown pentagrams) and VQDM (pink crosses). While seen samples are well separated, unseen forgeries often fall into low-confidence regions between the two clusters. As illustrated in Figure~\ref{fig:tsne}(a), many ADM samples lie within these regions (gray dashed ellipse), consistent with its lower best accuracy of 85.2\%\citep{jia2025secret}. VQDM samples show fewer overlaps (Figure~\ref{fig:tsne}(b)), aligning with its higher baseline of 96.8\%\citep{jia2025secret}. In contrast, mixup samples construct transitional samples that enrich low-confidence regions, improving coverage of unseen-source fakes and enhancing cross-source generalization, achieving a 15.44\% gain on ADM and 2.46\% on VQDM.

\begin{figure}[h]
  \centering
 
  {
        \includegraphics[width=1\linewidth]{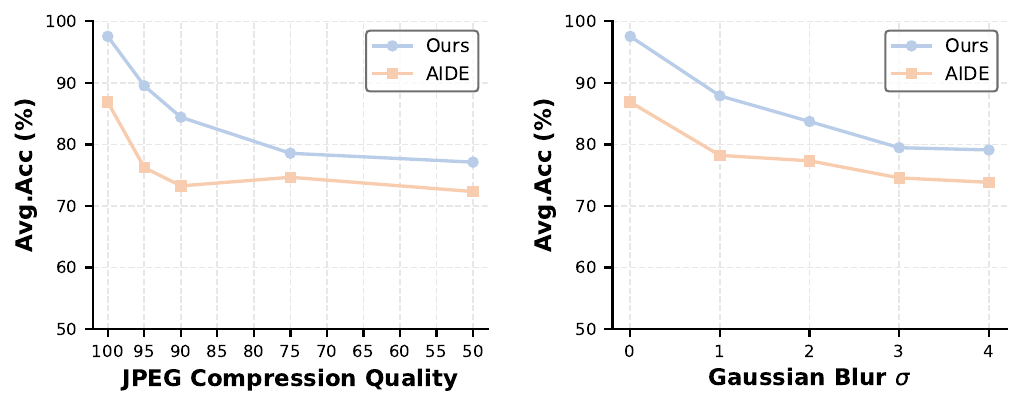}
     
     }
     % \vspace{-5pt}
   \caption{Robustness evaluation on GenImage dataset under JPEG compression and Gaussian blur.}
   \label{fig:robustness}
   \vspace{-10pt}
\end{figure}
\subsection{Robustness Evaluation}
% \noindent \textbf{Robustness Evaluation.}
To evaluate the robustness of HiMix under real-world degradations, we conduct experiments on the GenImage dataset by applying JPEG compression and Gaussian blur with varying intensities. These perturbations simulate common post-processing operations encountered in practical scenarios. As shown in \cref{fig:robustness}, although both degradations inevitably reduce overall performance due to the suppression of high-frequency forgery traces, HiMix consistently outperforms AIDE across all distortion levels. This indicates that HiMix achieves better resilience and maintains stable performance under both blurring and compression. We attribute this advantage to the mixup-driven augmentation, which encourages more stable decision boundaries under distributional shifts, and the hierarchical fusion mechanism, which enables the model to retain discriminative cues even when artifact patterns are significantly weakened. Overall, these results suggest that HiMix maintains reliable detection capability under common image degradations, highlighting its potential for real-world applications.

\subsection{Ablation Study} 
\noindent \textbf{Impact of each module.} As detailed in \cref{tab:ablation_combined}, we perform a hierarchical ablation study to validate the core components of HiMix. At the macro level (top half), the Mixup-driven Distributional Augmentation (MDA) module contributes the most significant gain (+8.49\% Avg.Acc) by effectively covering low-confidence regions between real and fake samples. The Hierarchical Artifact-aware Representation (HAR) module and the LoRA adapter further improve an average Acc by 5.30\% and 0.97\%, respectively. Combining all modules yields the optimal performance, verifying their strong synergy. At the micro level (bottom half), we dissect the internal design of the HAR module. While adding hierarchical region pooling (HiRP), cross-layer fusion (CLF), or cross-granularity fusion (CGF) to a CLS-only baseline provides incremental individual improvements, their hierarchical coupling delivers a substantial +5.30\% cumulative boost. This confirms that integrating multi-scale and cross-granularity cues is essential for robust artifact learning.

\begin{table}[t]
    \caption{Comprehensive ablation study of the proposed HiMix on the GenImage dataset. The table integrates both macro-level components (MDA, HAR, LoRA) and the micro-level architectural designs within the HAR module (HiRP, CLF, CGF). Average Acc (Avg.Acc) and average AP (Avg.AP) are reported in \%.}
    \centering
    \small
    \renewcommand{\arraystretch}{1}
    \setlength{\tabcolsep}{5pt}
    \scalebox{0.95}[0.95]{
    \begin{tabular}{cc|cccc|cc}
    \toprule
    \multirow{2}{*}{\textbf{\textit{MDA}}} & \multirow{2}{*}{\textbf{\textit{LoRA}}} & \multicolumn{4}{c|}{\textbf{\textit{HAR Modules}}} & \multirow{2}{*}{\textbf{\textit{Avg.Acc}}} & \multirow{2}{*}{\textbf{\textit{Avg.AP}}} \\
    \cmidrule{3-6}
    & & \textbf{\textit{CLS}} & \textbf{\textit{HiRP}} & \textbf{\textit{CLF}} & \textbf{\textit{CGF}} & & \\
    \midrule
    \multicolumn{8}{c}{\textit{Macro-level Component Ablation}} \\
    \midrule
    \ding{55} & \ding{51} & \ding{51} & \ding{55} & \ding{55} & \ding{55} & 83.69 & 94.69 \\
    \ding{55} & \ding{51} & \ding{51} & \ding{51} & \ding{51} & \ding{51} & 89.07 & 97.77 \\
    \ding{51} & \ding{55} & \ding{51} & \ding{51} & \ding{51} & \ding{51} & 96.59 & 99.69 \\
    \midrule
    \multicolumn{8}{c}{\textit{Micro-level HAR Ablation (with MDA \& LoRA)}} \\
    \midrule
    \ding{51} & \ding{51} & \ding{51} & \ding{55} & \ding{55} & \ding{55} & 92.26 & 97.59 \\ % S1
    \ding{51} & \ding{51} & \ding{51} & \ding{51} & \ding{55} & \ding{55} & 93.03 & 98.58 \\ % S2
    \ding{51} & \ding{51} & \ding{51} & \ding{51} & \ding{51} & \ding{55} & 95.72 & 99.57 \\ % S3
    \ding{51} & \ding{51} & \ding{51} & \ding{51} & \ding{55} & \ding{51} & 94.51 & 99.28 \\ % S4
    \midrule
    \ding{51} & \ding{51} & \ding{51} & \ding{51} & \ding{51} & \ding{51} & \textbf{97.56} & \textbf{99.71} \\ % Full Model
    \bottomrule
    \end{tabular}
    }
    \label{tab:ablation_combined}
    \vspace{-10pt}
\end{table}

\begin{figure}[h]
  \centering
  {
        \includegraphics[width=0.65\linewidth]{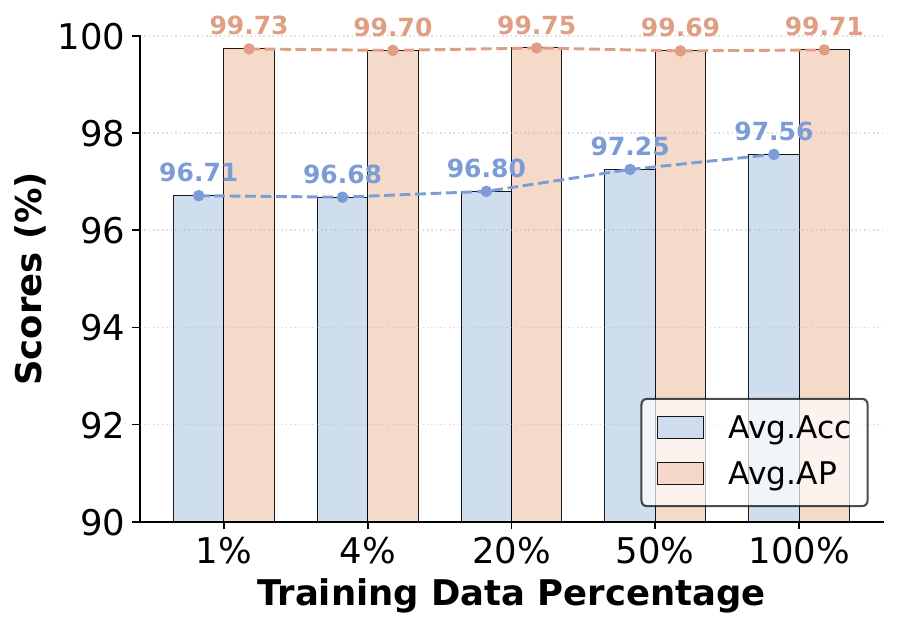}
     
     }
     \vspace{-8pt}
   \caption{Data efficiency evaluation of HiMix on GenImage dataset under varying training data size.}

   \label{fig:fewshot}
   \vspace{-3pt}
\end{figure}

\noindent \textbf{Impact of Different Training Data Size.}
We further examine the data efficiency of HiMix by varying the proportion of training samples (1\%, 4\%, 20\%, 50\%, and 100\%) from SDv1.4 and evaluate on GenImage. As shown in \cref{fig:fewshot}, with only 1\% of the data, HiMix achieves an average Acc of 96.71\% and an average AP of 99.73\%, a drop of just 0.84\% in the average Acc compared to the full dataset. This strong performance can be attributed to MDA, which expands the effective training distribution, and HAR, which captures generator-invariant artifact-aware representations, demonstrating strong data efficiency under limited training data.

\begin{figure}[h]
  \centering
  {
        \includegraphics[width=0.64\linewidth]{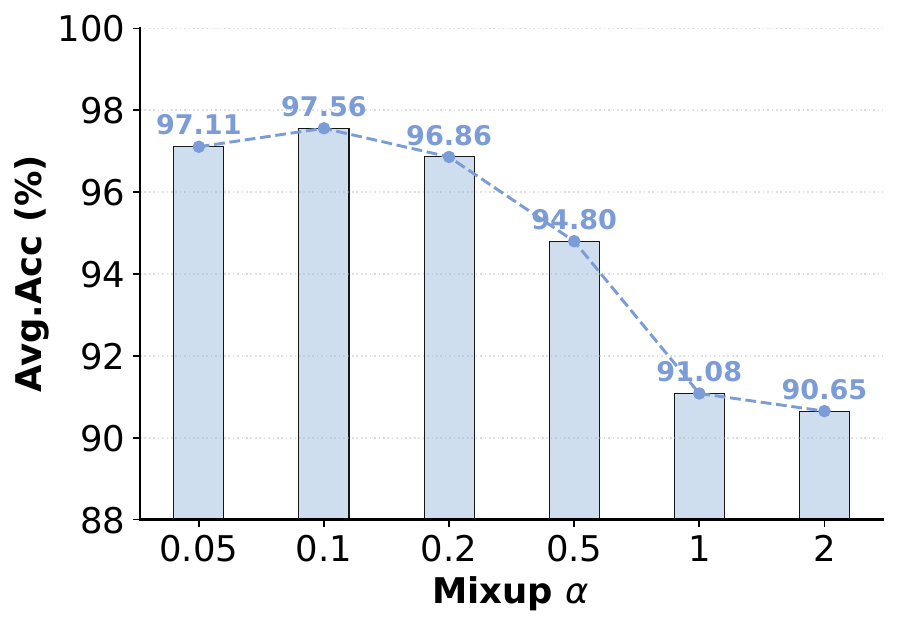}
     }
 \vspace{-8pt}
   \caption{Impact of the mixup coefficient $\alpha$ on the average Acc over GenImage dataset.}
   \label{fig:alpha}
   \vspace{-3pt}
\end{figure}

\noindent\textbf{Impact of Mixup Coefficient $\alpha$.} We study the effect of the Beta distribution coefficient $\alpha$, which governs the mixup interpolation degree. As shown in \cref{fig:alpha}, HiMix achieves a peak average accuracy of 97.56\% at $\alpha=0.1$. Smaller $\alpha$ values create near-real and near-fake boundary samples that enhance decision sharpness, while larger values ($\geq1$) produce overly smooth interpolations, blurring distinctive cues and reducing accuracy. This indicates that a bi-modal mixing strategy is essential for effective augmentation in forgery detection. \textit{Please refer to the supplementary material for more details on the mixup and LoRA ablation studies.}

\begin{table}[h]
\caption{Model complexity and inference efficiency.}
 \vspace{-8pt}
\small \centering
% \footnotesize
\renewcommand{\arraystretch}{1.15}
 \setlength{\tabcolsep}{4.5pt}
 \scalebox{0.95}[0.95]{
\begin{tabular}{lccc}
\toprule
\textbf{Method} 
& \textbf{Total Params (M)} 
& \textbf{Trainable (M)} 
& \textbf{FPS} \\
\midrule
CLIP baseline          & 428.67 & 1.05 & 105.38 \\
+ Hierarchical Pooling & 429.12 & 1.58 & 103.33 \\
+ Cross-layer Fusion   & 430.18 & 2.32 & 101.68 \\
+ LoRA \textbf{(Ours)} & 431.80 & 3.88 & 98.29 \\
\bottomrule
\end{tabular}}
\label{tab:efficiency_half}
 \vspace{-5pt}
\end{table}
\noindent \textbf{Computational Complexity.} \cref{tab:efficiency_half} reports the inference overhead of the proposed modules. Starting from the CLIP baseline (428.67M parameters, 105.38 FPS), hierarchical pooling and cross-layer fusion increase the model size to 430.18M parameters with a moderate drop of 101.68 FPS. With LoRA, the final model runs at 98.29 FPS and uses only 3.88M trainable parameters, indicating limited additional compute and memory overhead.

\section{Conclusion}\label{conclusion}
In this paper, we proposed \textit{HiMix}, a unified framework that improves the generalization of synthetic image detection. The Mixup-driven Distributional Augmentation (MDA) module constructs continuous real–fake transitions to expand the effective training distribution, with improved coverage of low-confidence regions, while the pixel-wise mixup operation reduces the influence of high-level semantics to promote artifact-aware learning. Meanwhile, the Hierarchical Artifact-aware Representation (HAR) module enables the model to capture complementary global and local cues, yielding more robust and discriminative feature learning under complex mixed-sample conditions. Extensive experiments across diverse benchmarks demonstrate the effectiveness of HiMix and its strong cross-generator generalization.

\begin{acks}
To Robert, for the bagels and explaining CMYK and color spaces.
\end{acks}

\bibliographystyle{ACM-Reference-Format}
\bibliography{sample-base}

% \appendix









\end{document}